# Deep Learning Neural Networks for Emotion Classification from Text: Enhanced Leaky Rectified Linear Unit Activation and Weighted Loss

**Hui Yang[1], Abeer Alsadoon[1,2,3,4*], P.W.C. Prasad[1,2,3], Thair Al-Dala'in[1,2,3,4], Tarik A. Rashid[5], Angelika Maag [1], Omar Hisham Alsadoon[6]**

[1] School of Computing Mathematics and Engineering, Charles Sturt University (CSU), Australia
[2] School of Computer Data and Mathematical Sciences, Western Sydney University (WSU), Sydney, Australia
[3] Kent Institute Australia, Sydney, Australia
[4] Asia Pacific International College (APIC), Sydney, Australia
[5] Computer Science and Engineering, University of Kurdistan Hewler, Erbil, KRG, IRAQ
[6] Department of Islamic Sciences, Al Iraqia University, Baghdad, Iraq

Abeer Alsadoon[1*]
* Corresponding author. Dr Abeer Alsadoon, [1] School of Computing Mathematics and Engineering, Charles Sturt University (CSU), Australia , Email: alsadoon.abeer@gmail.com , Phone +61 413971627

## Abstract

Accurate emotion classification for online reviews is vital for business organizations to gain deeper insights into markets. Although deep learning has been successfully implemented in this area, accuracy and processing time are still major problems preventing it from reaching its full potential. This paper proposes an Enhanced Leaky Rectified Linear Unit activation and Weighted Loss (ELReLUWL) algorithm for enhanced text emotion classification and faster parameter convergence speed. This algorithm includes the definition of the inflection point and the slope for inputs on the left side of the inflection point to avoid gradient saturation. It also considers the weight of samples belonging to each class to compensate for the influence of data imbalance. Convolutional Neural Network (CNN) combined with the proposed algorithm to increase the classification accuracy and decrease the processing time by eliminating the gradient saturation problem and minimizing the negative effect of data imbalance, demonstrated on a binary sentiment problem. All work was carried out using supervised deep learning. The results for accuracy and processing time are obtained by using different datasets and different review types. The results show that the proposed solution achieves better classification performance in different data scenarios and different review types. The proposed model takes less convergence time to achieve model optimization with seven epochs against the current convergence time of 11.5 epochs on average. The proposed solution improves accuracy and reduces the processing time of text emotion classification. The solution provides an average class accuracy of 96.63% against a current average accuracy of 91.56%. It also provides a processing time of 23.3 milliseconds compared to the current average processing time of 33.2 milliseconds. Finally, this study solves the issues of gradient saturation and data imbalance. It enhances overall average class accuracy and decreases processing time.

## 1. Introduction

With the rapid development of e-commerce and social networks, the Internet contains more and more user-generated reviews [1-3] consisting of user opinions and emotions, which have significant business implications [4, 5]. Because the emotions reveal user attitudes towards a product or service, they can be used for predicting sales, stock market trends, and more [6]. Therefore, research on emotion classification techniques of texts has significant commercial value for companies [7, 8]. A short example from a movie review illustrates some of the problems:

Charlie's      Wilson's     War demonstrates with deft veracity just how futile wars can be, especially to the very people who spend countless hours and finances to fund them.

Negative

Potentially negative; may be sarcasm since weapons sales are lucrative

Sample Source: IMDB https://www.kaggle.com/lakshmi25npathi/imdb-dataset-of-50k-movie-reviews

This means that the classifier of the overall emotions must be capable of detecting sarcasm and balancing binary elements in a longer review to arrive at an overall conclusion. Traditional methods for classifying emotions expressed in a text can be categorized into the lexicon-based and machine learning-based methods [9]. The former methods use lexicons to calculate an emotion score for the input sentence to produce classification results [9]. It depends heavily on the quality and the comprehensiveness of the lexicon [10]. The latter method employs hand-crafted feature engineering to capture statistical features (e.g., presence of frequent n-grams) to create the representations of input sentences [11, 12]. These representations are then fed into classifiers, such as support vector machines (SVM) to output the prediction [13]. Manual feature engineering is time-consuming as each feature must be built one at a time resulting in low performance for classification [14]. Furthermore, the code is linked to individual problems and is not transferable to new datasets. Emerging deep learning methods with the ability to extract features automatically can generally solve these issues, although this is less certain when domain knowledge is needed.

Notwithstanding such issues, deep learning methods have proven their effectiveness in the field of natural language processing (NLP) [15, 16]. Convolutional neural networks (CNN) are effective deep learning methods used to capture features from sentences automatically [16]. They can be supervised or unsupervised. This work has been carried out using supervised CNN. Compared with other deep learning methods, CNN can better extract local features and have a lower computational cost [17]. CNN has been successfully adopted in solving NLP problems, especially emotion analysis [18]. The architecture comprises input, convolutional, pooling, and fully connected layers with classifiers to learn and extract features from sentences and produce emotion class probabilities [19]. However, the overall performance of CNN models still needs improvement due to the influence of parameters and activation and loss functions. Optimized algorithms are needed to enhance average class accuracy and processing time of CNN models for emotion classification.

Current CNN models for text emotion classification use various techniques and algorithms to increase average class accuracy and lower processing time. The work of each is valuable in its own way, and some have achieved astounding classification accuracies, e.g. 95.7% [20]. However, this was generally at the expense of generalizability as either overfitting, gradient saturation, computational cost, sample sizes or a combination of these remain unresolved. Furthermore, all researchers have trained and tested their systems on different datasets. Therefore, performance has been stated in absolute terms only, due to the problem of comparing numerical outputs when the input is variable, which prevents comparison. In the case of Liu [21], they compared their innovation across two databases only. However, since all three systems were tested on the same two databases, it can be stated that, for these two databases, the results are conclusive.

Liu's [21] experimental work (CBOW+D-CNN) with COAE2014 and IMDB datasets produced 87 and 90% classification results when their dropout strategy was introduced into the Softmax classifier of the CNN. This comparison was made across the two databases using the same methodology but three different deep learning configurations (i.e., N-Gram+CNN; CBOW+CNN; CBOW+D-CNN). Whilst the classification results were impressive, the more interest is here the dropout algorithm that reduces overfitting problems and a CNN with this function, based on two databases, performs better than the traditional models. This gives high importance to this work as overfitting has been a troublesome issue for some time. This has prompted us to select this study as the Baseline State-of-the-Art (SOTA). We selected Liu's [21] study even though it included a sigmoid activation function which can cause gradient saturation problems [22]. Furthermore, it was impossible to ascertain the extent to which accuracy might have been negatively affected by imbalanced data since Liu [21] did not address this issue.

This paper proposes an Enhanced Leaky Rectified Linear Unit activation and Weighted Loss (ELReLUWL) algorithm for enhanced text emotion classification and faster parameter convergence speed. It defines the inflection point and the slope for input values as smaller than the inflection point for the activation, which stops the gradient from saturating and is less computationally complex. It also defines each class's weight when backpropagating the loss, which compensates for the negative effect of unbalanced classes in the dataset. The algorithm significantly impacted the average class accuracy for emotion classification in user generated online reviews and reduced the processing time by combining a modified leaky rectified linear unit function with a modified loss function. This study, which addresses a binary sentiment problem, aims to reduce the risk of gradient saturation and reduce the negative effect of the imbalance in the training data, thus improving the classification ability of the system and accelerating the parameter convergence. Training data imbalance is common in any form of sentiment analysis and, more so, for emotions, where a wide range of different classes of emotion may lead to "skewed training data" [23].

The structure of the remainder of this paper is as follows. Section 2 is a literature review that outlines current methods in text emotion classification using deep learning. Section 3 presents our proposed solution while the experiment setup and the results are illustrated in Section 4. The discussion is presented in Section 5. Finally, we conclude this paper in Section 6, which summarizes contributions and limitations and suggests directions for future work.

## 2. Literature Review

Deep learning systems, which do not rely on extensive hand-crafted feature engineering [14], have attracted researchers' attention in NLP for some time as they can automatically extract features. A range of these techniques used for text-based emotion classification is here reviewed. These models are mainly based on three types of neural networks: Convolutional Neural Networks (CNN), Long Short-Term Memory (LSTM), and a Hybrid of CNN and LSTM. The baseline SOTA is presented at the end of this section.

### 2.1 Emotion classification based on CNN

CNN is frequently selected for emotion classification and have achieved excellent results in a range of studies. Stojanovski et al. [14] proposed a model to improve sentiment analysis emotion classification. They proposed a model that combined CNN with GloVe word embedding to obtain enhanced accuracy. They included a sigmoid activated layer and a tangent activated layer in the CNN to improve performance further. They also suggested that adding more layers to the network would improve performance. This research achieved considerable improvements with an accuracy of 58.84 in emotion identification and an F1 score of 64.88 in sentiment analysis. However, the training data imbalance problem still needs to be solved [24]. Zhang et al. [9] improved sentiment classification in a text by proposing a model using three different word embeddings to integrate the sentiment meaning into semantic embeddings. The coordination of the attention vector with CNN enables the model to extract both local and global features.

Additionally, cross-modality consistent regression (CCR) further enhances the model's performance. It achieved remarkable high performance with 0.6% improvement in the F1-score and an accuracy of 88.83. However, the overfitting issue that commonly occurs in the deep neural network was ignored, which decreases overall average class accuracy [25]. Moreover, multiple layers increase computational cost and processing time.

Liu [21] solved the overfitting problem and provided improved emotion classification from texts. He adopted a dropout strategy in the softmax classifier which significantly reduced the overfitting problem commonly occurring in traditional CCN methods. He further utilized a Continuous Bag of Words (CBOW) language model for the CCN to improve emotional information capture. The solution provides robust and precise text emotion classification in both English and Chinese, with an accuracy of 87.2% and 90.5%, respectively. However, the sigmoid function used in the convolutional layer is prone to gradient saturation which, as argued by Zhao et al. [26] in their evaluation of the sigmoid function, decreases the overall performance. In addition, there is no strategy for dealing with the imbalanced data, which is commonly encountered in this domain [24].

Phillips et al. [20] solved the gradient saturation problem and increased the average class accuracy with an 8-layer CNN model with optimized structure. After experimenting with different functions, they selected a leaky rectified linear unit as the activation function and 'max-pooling' as the pooling function. This work achieved outstanding performance with an accuracy of 97.65%, sensitivity of 97.96%, and specificity of 97.35%, respectively. However, the multiple layer structure increases the computational complexity [27] and is not efficient in handling small-size datasets. Thus, the hyperparameters need to be optimized through random search methods rather than only by experience [20].

Gu et al. [28] improved the performance of aspect-based sentiment classification with a cascaded model with 2- level CNN. The proposed cascaded CNN contains aspect mappers to map the input sentence into the corresponding aspect accurately. When the input belongs to predefined aspects, the sentiment classier will determine the sentiment polarity. The pre-trained word2vec embeddings significantly improve the performance of aspect mapping and emotion classification. This solution outperforms traditional machine learning approaches with F1 score up to 83.74 in aspect mapping, and an accuracy of 84.87% in sentiment classification. However, the system contains only a single emotion classifier, which is not suitable in complex textual environments [29]. When encountering more than one emotion type in a sentence, the classification will not be accurate. Zhang et al. [2] proposed a model to enhance text emotion classification. They proposed a 3-way-CNN with a confidence divider to separate the test data into three classes (i.e. negative, positive and boundary). A combination of Support Vector Machine and Naive Bayes (NB-SVM) is used as the enhanced model to reclassify data in the boundary region. Although the solution provides improved accuracy at 90.3% in text emotion classification, its performance is restricted by CNN and NB-SVM. If the test data is difficult to classify for both CNN and the enhanced model, this model will not be able to provide an accurate prediction [2]. Additionally, although the confidence function is essential for this model to effectively classify the boundary data, it still needs optimization to achieve further improvements.

## 2.2 Emotion classification based on Long Short-Term Memory (LSTM)

Unlike CNN, LSTM, a variation of Recurrent Neural Networks (RNN) [30], can identify long-term dependencies in the input [31, 32]. Different models using LSTM have been proposed to improve the capability of emotion classification. Mahmoudi et al. [33] compared a range of approaches from traditional machine learning methods to deep learning approaches for improved emotion prediction in the stock market. Then they introduced a solution using LSTM to seize long-term dependencies to emphasize the linguistic objects. They adopted domain-specific word embedding GloVeST with non-static capabilities that allow manipulation of word positions to reflect the word's positive or negative character more accurately. The solution achieved enhanced investor emotion classification performance, as confirmed by the Wilcoxon Sum-Rank Test (WSuRT) with p≤0.0283. However, the researchers identified the limits of their work as originating from the small size of the data set as they had to rely on 'StockTwits' alone. Furthermore, this research classification is limited to only two types (i.e. bullish and bearish) which becomes a problem in a bullish market where the training data class imbalance is of such magnitude that it requires extensive modifications before it can be applied to other domains [6].

Chatterjee et al. [34] improved deep learning baseline approaches to predict emotions from user utterances. They proposed a solution with two LSTM layers with semantic and sentiment features for more precise emotion classifications. This significantly improved emotion prediction over other solutions with p<0.005 in the McNemar's test. However, the conversation context is not taken into consideration, which may lead to classification errors. Furthermore, accuracy can be further improved by extending the research to more emotion classes such as fear, surprise, and so on [6]. Rao et al. [35] improved the accuracy of document-level emotion classification. They proposed a model which contains two LSTMs to extract enough successive document representations from a document. The first LSTM can capture the semantic representation of sentences; while with the second LSTM layer, the model can extract the associations of sentences in a document. It achieves average class accuracy of 0.639 and MSE of 0.46, respectively.

However, setting a maximum number of sentences in a document limits performance [36]. Also, this work models documents sequentially, which is not suitable for domains where documents are not composed in a sequential way. Kratzwald et al. [24] introduced several remedies to improve the performance of inferring the emotional state within narrative contents. The proposed customized LSTM model integrates dropout regularization, bidirectional processing, and weighted loss functions to deal with the data imbalance problem. The function loss can significantly reduce using regularization [3, 37]. It outperforms the other options by 23.2% for the F1- score and 11.6% for MSE, respectively. Also, sent2affect implementation of transfer learning further enhances improvements by up to 6.6%. However, this area's datasets are still exceedingly small compared to other deep learning applications [6]. Thus, larger datasets need to be created to take full advantage of deep learning.

## 2.3  Emotion classification based on a Hybrid of CNN and LSTM

Many approaches combine CNN and LSTM models to capture both short-range and long-range dependencies [29], and in many cases, these combinations have provided SOTA systems. Chen et al. [36] improved accuracy in sentence-level sentiment classification by proposing a model containing a bidirectional LSTM and conditional random fields (BiLSTM-CRF) to classify the sentence type. The model also includes a 1- dimensional CNN to classify the emotion, tailored to specific sentence types by identifying the number of emotions expressed in each of the sentences. By dividing these sentences into groups, it reduces the complexity for each category which means, the learning algorithm has less complexity to deal with in the training data. This pipeline strategy obtains improved accuracy with the binary labeled version of Stanford sentiment treebank (SST) (88.3%) compared to 85.4% for Customer Reviews (CR). However, since in the SST neutral sentences have been removed, a real comparison is limited. This work is also limited in terms of sentence-level analysis [35] as the maximum sentence length is set and sentences exceeding this length are shortened while sentences that do meet the minimum requirements are lengthened by 0 vectors. The CNN will not be able to generate enough relationship information between sentences in a long text [17]. Also, the fixed maximum length of a sentence needs optimization to avoid losing useful information. Lin et al. [38] achieved enhanced emotion prediction in multi-media. They introduced an Explicit Emotion Signal based emotion extraction method to catch the middle-level features from images and texts. This solution uses a network to train each modality (LSTM for text and CNN for images) and then combines the features. It achieves an acceptable performance (accuracy of 56.4% in text sentiment learning) while only requiring 3% training samples for the Visual Geometry Group (VGG) model in images and 43% training samples for the LSTM model in texts, to achieve the same performance. However, a multi-label learning problem exists, which slows down processing [38].

Sun and Zhang [30] improved the accuracy of emotion classification from social media. They introduced a hybrid method combining CNN and LSTM. It also involves a multivariable Gaussian method for multi-dimensional emotion processing and enables better anomaly detection. It obtains an average class accuracy of 94.19%, abnormal emotion detection accuracies of 83.49% (in terms of the individual user) and 87.84% (in terms of the different month). It considers the local semantic information (which is extracted by CNN) and takes global contextual information (which is done by the LSTM channel) into account. However, selecting thresholds for abnormal emotion is time-consuming, which slows down the system [34]. Therefore, a solution needs to be found to shorten the computation time for selecting the optimal threshold for a more efficient model. Colnerič and Demšar [6] enhanced emotion prediction with three emotion classification systems (i.e. POMS's, Plutchik's and Ekman's classifications of emotions) at one time. They proposed a unison model which works on a large amount of data while using a weighted sampling strategy for more precise results. This unison model can predict three classifications at once and achieve comparable performance in emotion prediction over single models, although the class imbalance problem remains unsolved [24]. Therefore, relations between emotion classifications need to be explored to expand the prediction to other classifications and develop a universal emotion detection algorithm that is not constrained to one domain.

Sun [39] advanced the performance of short text emotion classification. They proposed a hybrid method to extract local features and global context information to improve emotion classification accuracy. They also used multi-granularity data augmentation [40] to pre-train the hybrid model to create sufficient data representation for enhanced classification ability. This solution shows great improvements in short term emotion classification, with F-score reaching 0.9054. The combination provides a model with more in-depth insight into data, despite data imbalance problems [24], and unsupervised pre-training [27].

Clearly, there are numerous options for selecting a deep learning algorithm, depending on the intended application. However, there remain challenges in the form of overfitting, gradient saturation, data imbalance, small data sets and dropout regularization. From among the above research, we considered Liu [21], Phillips et al. [20], and Vinay Kumar et al. [4] for our study. Liu [21] was considered since they solved the problem of overfitting through the addition of a dropout strategy in the softmax classifier. Phillips et al. [20] was considered because they used a weighted loss function to deal with data imbalance, and Vinay Kumar et al. [4] was selected because they addressed dropout regularization. From among these, we chose Liu [21] as our base paper because it is built around CNN, which allows larger data sizes. Vinay Kumar et al. [4] also used CNN, but the multiple layer structure they built increased computational complexity, and their search method was based on experience only to which randomization needs to be added to optimize results. Whilst we will draw on elements of Vinay Kumar et al. [4] for our work, it is not suitable as a base paper because it uses LSTM which cannot process large data sets. The following section explore Liu [21] work as it's our baseline SOTA.

## 2.4 Baseline SOTA

This section describes our base model's features and techniques for text emotion classification, proposed by Liu [21]. As shown in Fig.1, useful features are highlighted inside the broken blue line, and the limitations are highlighted inside the broken red line. Liu proposed Dropout regularized CNN combined with Continuous Bag of Words (D-CNN+CBOW) to enhance text emotion average class accuracy. This solution applies a dropout strategy to the softmax classifier (highlighted in blue in Fig.1) to effectively prevent overfitting and enhance generalization ability [21]. Therefore, the overall average class accuracy has improved. CNN can automatically extract semantic features from the text, without a large amount of manually labelled data. The CBOW model continuously constructs distributed word representations for CNN to capture emotional information more effectively. Moreover, the combination of CBOW and D-CNN increases convergence speed. Compared with traditional methods, this solution provides more robust and more precise text emotion classification in both English and Chinese, with an accuracy of 87.2% and 90.5%, respectively. This model has two main stages (see Fig.1): 1) CBOW language model for constructing word vector representation; 2) Dropout regularized CNN for feature extraction and classification.

*CBOW model constructs word representations*: This includes input, projection, and output layers. The input layer first reads the corpus words sequentially and maps them in the projection layer. The index of each word is obtained using the hash table. The projection layer calculates the cumulative sum of surrounding context words for the target word. After that, the output layer performs regression analysis operation on the cumulative sum of context to generate the target word's vector value. Unlike traditional word bag models, this system generates continuous distributed representations of words, which forms more suitable word embeddings for DNN to capture text features more effectively [41].

*Dropout regularized CNN for feature extraction and classification*: This consists of the convolutional layer, pooling layer, and a fully connected layer. The CNN takes the word vectors obtained from the CBOW model as input, then transforms them into a matrix, where the number of rows is the sentence length, and the number of columns is the word vector dimension [42]. The filters (sizes of 2, 3, and 4, each size has 2 filters) in the convolutional layer convolve upon the matrix to create the feature maps. Then the feature maps are condensed by the pooling layer to decrease the number of parameters. The maximum value of each map is extracted, forming a single feature vector as the next layer's input, which captures the most important information. After that, the dropout algorithm is applied in the softmax layer by randomly disabling parameters according to a predefined ratio [43]. This dropout can effectively minimize the possibility of overfitting and increase CNN's generalization capability [44]. Simultaneously, the classification result is produced by using the softmax function with the final parameters of the network. In addition, this training process also adopts a backpropagation strategy, which distributes the overall error backwards and updates the weights and biases through a random gradient. The advantage of backpropagation is that it optimizes the network settings and minimizes errors [21, 24].

The experiment results illustrate that this solution has achieved remarkable text emotion average class accuracy, with 90.5% in Chinese and 87.2% in English, respectively. However, the sigmoid function is used as the activation function in a convolutional layer, which is likely to be affected by a gradient saturation problem in gradient-based learning models [26]. Another limitation is that there is no strategy for handling the imbalanced data, which is commonly involved in natural language processing. The gradient of the sigmoid function tends to vanish when the function value approaches 0 or 1. It causes the neurons to backpropagate the gradient approaching 0, which means the neural network fails to update the parameters to optimize the training [45]. Therefore, the use of the sigmoid function in convolutional layers impacts the model's recognition ability and reduces overall class accuracy. It also slows down convergence, thus making the model less efficient. Furthermore, without handling imbalanced training data, the model is prone to bias towards frequent training data classes [14, 24]. This bias causes inaccurate results and leads to a decrease in classification performance. This solution provides a precise text emotion classification in both English and Chinese, with an accuracy of 87.2% and 90.5%, respectively. The whole training algorithm is presented in Table 1, while the flowchart is presented in Fig. 2.

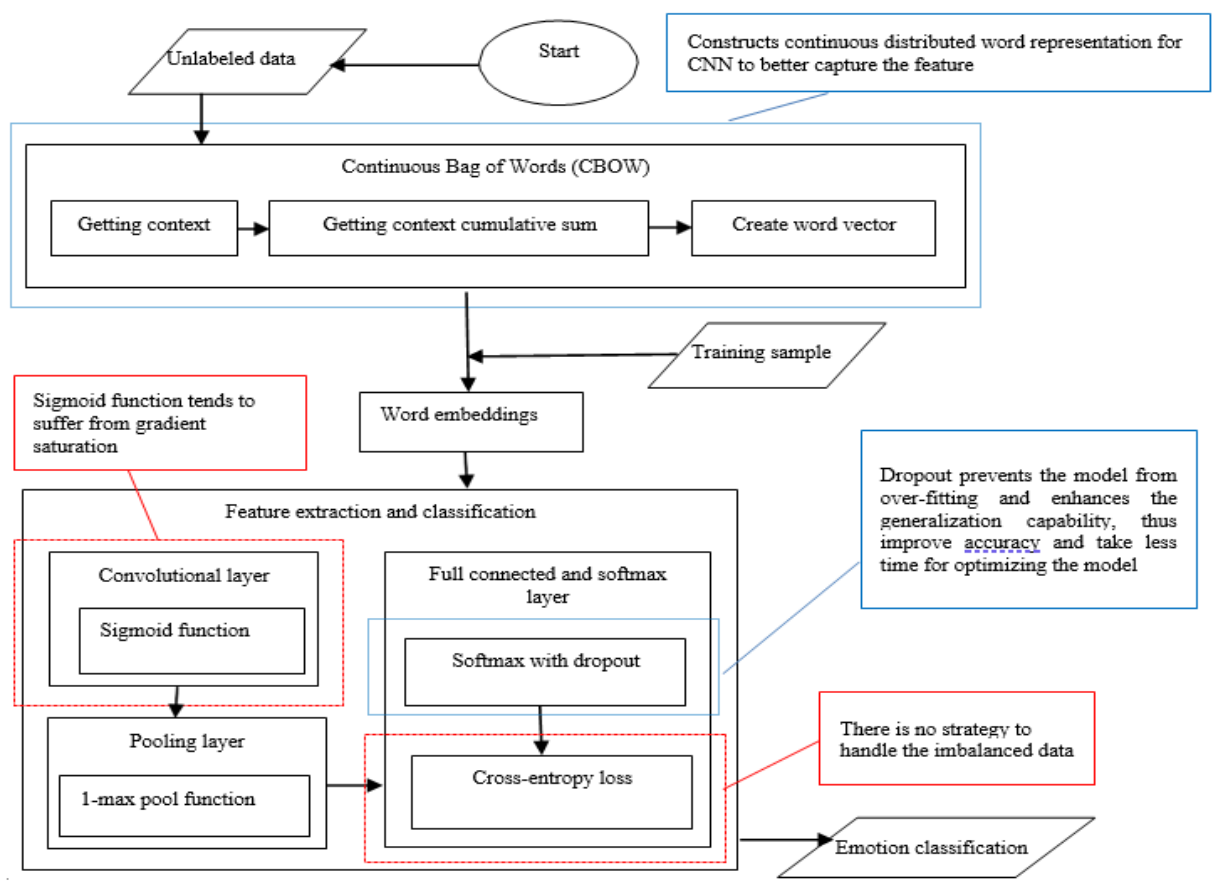

Fig.1: workflow of the Baseline SOTA [21]
(The blue border illustrates the good features and the red border illustrates the limitations)

In the convolutional layer, the output of each neuron is calculated by the activation function. It uses the weight and bias and the preceding layer's input to collect the features from the text. In this solution, the sigmoid function is used to compute neurons' output, which is expressed in equation (1) [21]. And the input of the sigmoid function is generated with the input of the previous layer, weight, and bias, which is expressed in equation (2) [21]. However, the sigmoid function is prone to suffer from gradient saturation, which affects the average class accuracy and reduces efficiency. This problem can be solved using the leaky rectified linear unit (LReLU) function [20].

$$u_j = \frac{1}{1 + e^{-x}} \hspace{4cm} (1)$$

Where,

$j$: the $j^{\text{th}}$ feature map

$e$: the natural logarithm

$u_j$: the activation function

$x$: the input, which can be expressed in equation (2):

$$x = \sum_{i=1}^{n} u_i^{l-1} * w_{ji} + b_j^l \hspace{3cm} (2)$$

Where,

i: the $i^{\text{th}}$ neuron

n: the total number of input neurons

l: the $l^{\text{th}}$ layer

j: the $j^{\text{th}}$ feature map

$u_j^{l-1}$: the input neuron from the $l$-$1^{\text{th}}$ layer for the $j^{\text{th}}$ feature map

$w_{ji}$: the weight of the $i$th neuron for the $j^{\text{th}}$ feature map

$b_j^l$: the bias at the $l$th layer for the $j^{\text{th}}$ feature map

The goal is to minimize the cross-entropy loss [46], which can be defined in equation (3). However, the imbalanced data problem has not been considered, which can impact the overall average class accuracy. This problem can be solved by calculating the weighted loss, which multiplies the loss of each input with a corresponding weight [24].

$$L = -\sum_k \sum_n d_k \log P_k \hspace{4cm} (3)$$

Where,

k: the numbers of the classes

n: the numbers of samples

L: cross-entropy loss

$d_k$: the desired output of class $k$

$P_k$: the predicted probability of class $k$

Table 1: D-CNN+CBOW algorithm

Algorithm: D-CNN+CBOW for text emotion classification

Input: Unlabelled text $D_U$, training data $D_T$, test data $D_T$

Output: The emotion class label of the evaluation text

BEGIN

Step 1: For each sentence $s$ in $D_U$

For each word $w_t$ in $s$

Get context words of the target word

Calculate $\log p(w_t | w_{t-k}, \ldots, w_{t+k})$

End For

Maximize the objective function: $\frac{1}{n} \sum_{n=k}^{n-k} \log p(w_t | w_{t-k}, \ldots, w_{t+k})$

End For

Export a collection of word vectors of all the words in $D_U$: $W$

Step 2: For each sentence $s$ in $D_T$

Look up the word vectors of all the words from $W$

Carry out the convolution on the vectors to extract the feature values

Combine all the feature values into feature vectors

Perform the pooling operation to extract the most important features

End For

Export trained CNN model and feature vectors of training data set

Step 3: For each sentence in $D_T$

The feature vectors of evaluation sentences are obtained by the trained CNN model

Classification by Dropout regularized Softmax classifier

END

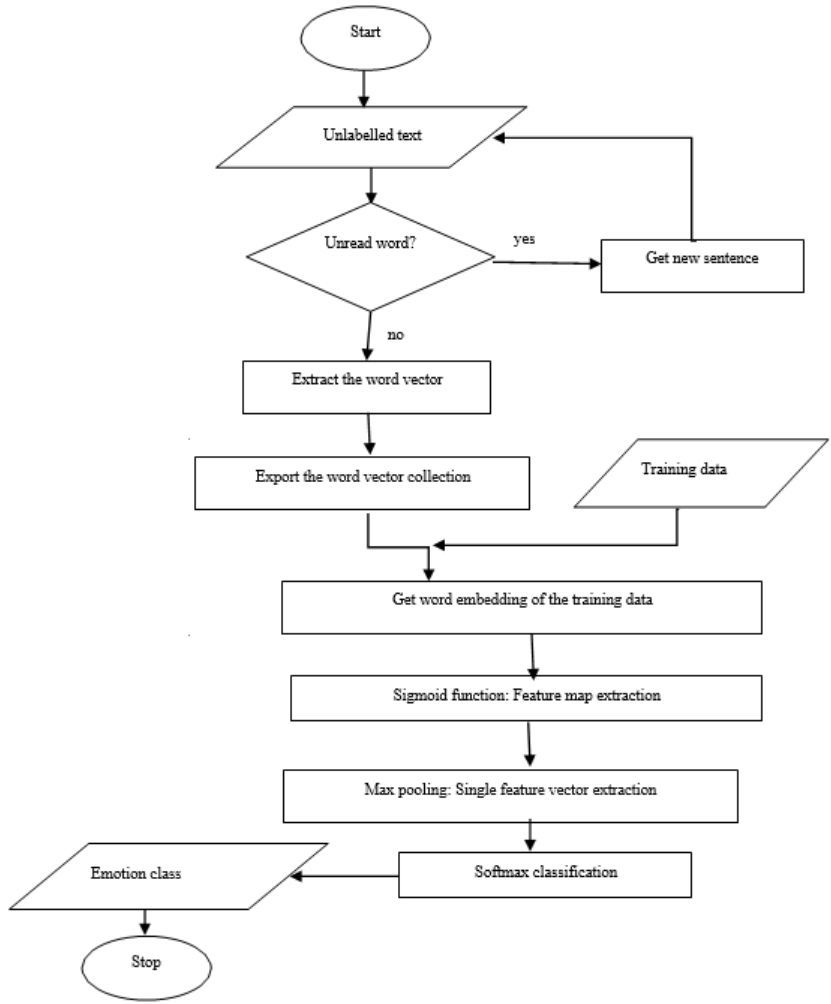

Fig.2: Flowchart of State-of-the-Art model

# 3. Proposed System

After reviewing various feature extraction and classification methods using deep learning for text emotion classification, each method's advantages and drawbacks are analyzed. It was found that the activation function, loss function, and weight are the paramount factors that affect CNN's performance.

From the list of reviewed classification methods, the solution proposed by Liu [21] is selected as the baseline SOTA work and used as the foundation for the proposed solution. The main reason for selecting this solution is to propose a Dropout regularized Convolutional Neural Network (D-CNN) to prevent overfitting. Because of overfitting, it is difficult for the neural network to generalize the unknown data, which results in poor performance [35]. Dropout randomly selects parameters participating in the training. Therefore, the neural network can be turned into multiple combinations to enhance generalization ability effectively and avoid overfitting, thus improving average class accuracy [21]. In addition, the D-CNN is combined with a Continuous Bag of Words (CBOW) model, which further enhances overall average class accuracy. The CBOW model constructs continuous distributed word representations which enable CNN to capture emotional information more effectively.

However, there still are some limitations to this work. The activation function is the sigmoid function, which is likely to cause gradient saturation problems [20]. The gradient saturation impacts the overall average class accuracy and slows down the convergence speed [45]. Another limitation is that there is no strategy for handling the imbalanced data, resulting in a bias towards more frequent classes in the training data [24]. While Liu [21] used the same data sets for testing and training, there were some inconsistencies in terms of corpus size and data split. The IMDB data set is perfectly balanced, although ten times larger than the COAE2014 for which data are skewed towards the positive by 42%.

This bias leads to inaccurate results and to decrease the training data classification performance. To overcome the gradient saturation problem, the proposed solution introduces a modified rectified linear unit (MLReLU) function inspired by Phillips et al. [20]. They draw their insights about RELU from their own 2015 work in which they relied on Hara et al. [47] to justify the use of this tool. They have argued that RELU is a generic non-linear function that merely facilitates passing data from one deep learning layer to the next and that currently, RELU is the choice for many applications. They also point to its versatility by referring to the work of Microsoft, IBM, Yahoo!, Twitter, and Adobe who work with RELU and RELU versions for text extraction.

Another modified feature of this solution is calculating the weighted loss adapted from the work by Kratzwald et al. [24], to deal with the imbalanced training data. These two enhanced features solve some of the current solution's problems and improve average class accuracy and convergence speed. The proposed system consists of two main stages (Fig.3): (1) word representation and (2) feature extraction and classification.

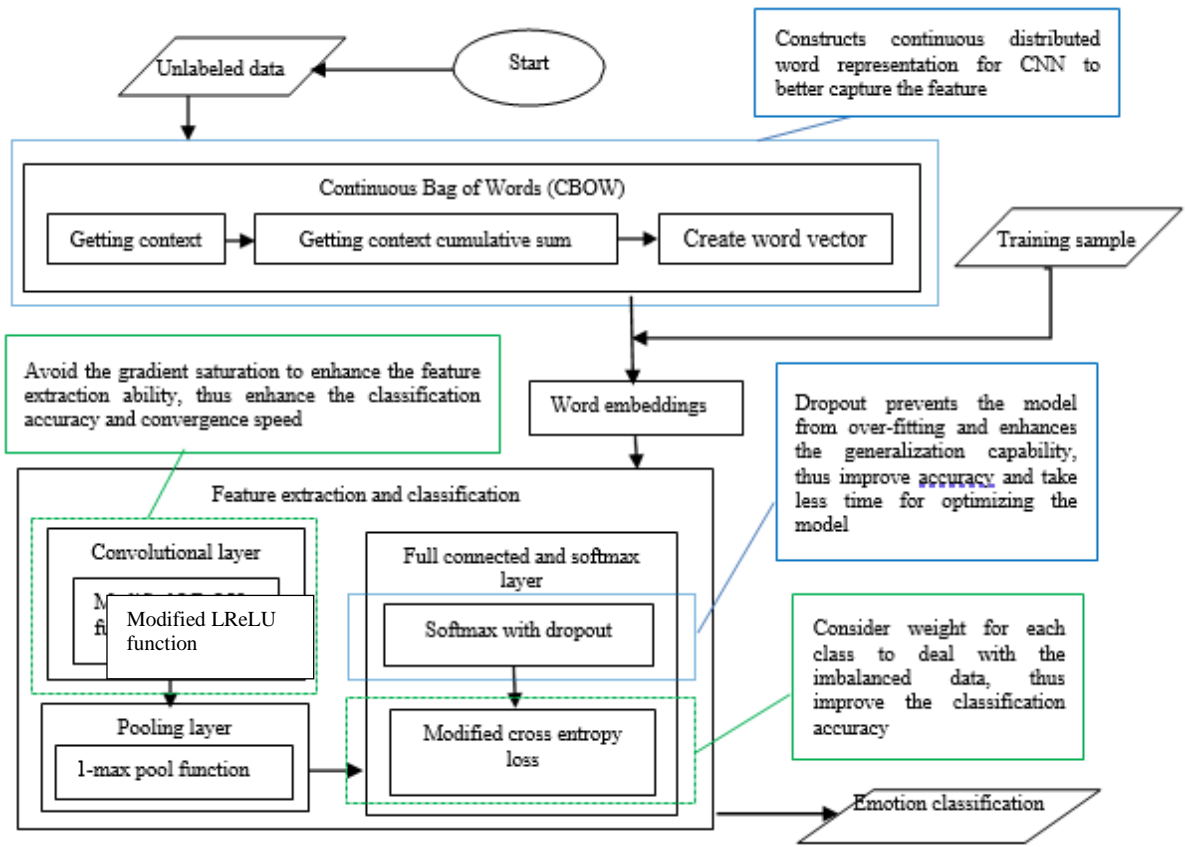

Fig.3: The workflow of the proposed model for emotion classification using LReLU activation and modified loss function (the green borders refer to the new components in the proposed model)

*Word representation*: follows the continuous bag-of-words architecture by Liu [21] which consists of the input, projection, and output layer. The input layer first reads the corpus words sequentially and maps them in the projection layer. Then the index of each word is obtained using the hash table as shown in Fig.3. The projection layer calculates the cumulative sum of context words for the target word. After that, the output layer performs a regression analysis operation on the cumulative context sum to generate a vector value for the target word.

*Feature extraction and classification using CNN*: The CNN model consists of a convolutional layer, a pooling layer, and a fully connected layer. The CNN takes training samples as input and maps them into word vectors according to the CBOW word representation collection. The training sample is then converted into a matrix where the number of rows corresponding to the sentence length and the columns are the word vector dimensions. The filters in the convolutional layer convolve on the matrix to create a variable feature map. The activation function calculates each unit in the convolutional layer. Here the modified LReLU (MLReLU) function is used. Compared to the sigmoid function, MLReLU is simpler and solves the gradient saturation problem. It allows the neurons to update effectively and accelerates the convergence of the CNN model [20]. After that, the feature maps are refined by the pooling layer to extract each map's maximum value. Then all the maximum values form a single feature vector, which contains the most relevant information. And then the model randomly drops out parameters at a predefined rate. After dropout, the classification result is produced by using the 'softmax' function with the network's final parameters. Then the cross-entropy loss is calculated using a modified cross-entropy loss function. The modified loss function takes each class's weight into account and reduces the bias towards the more common classes [24]. After that, the calculated loss is backpropagated through the neural network to update the weights and bias for optimization.

The complete training strategy Enhanced Leaky Rectified Linear Unit activation and Weighted Loss (ELReLUWL) algorithm is presented in Table 2, and the flowchart of this algorithm is presented in Fig. 4.

Table 2: Enhanced Leaky Rectified Linear Unit activation and Weighted Loss (ELReLUWL) algorithm

Algorithm: ELReLUWL for text emotion classification
Input: training data $D_T$, validation data $D_V$
Output: The emotion class label of the text
BEGIN
Step 1: For each sample $s$ in $D_T$:
        Convert all the words in $s$ into word vectors
Step 2: Compute the convolutional outputs to extract the feature values using MLReLU activation
Step 3: Combine all the feature values into feature vectors
Step 4: Perform the max (feature vectors) to extract the most important features
Step 5: Perform dropout on the feature vectors
Step 6: Calculate the probabilities for each class using softmax function
Step 7: Compute the error using the modified loss function
Step 8: Backpropagate the error through the model to update the parameters of hidden layers
END

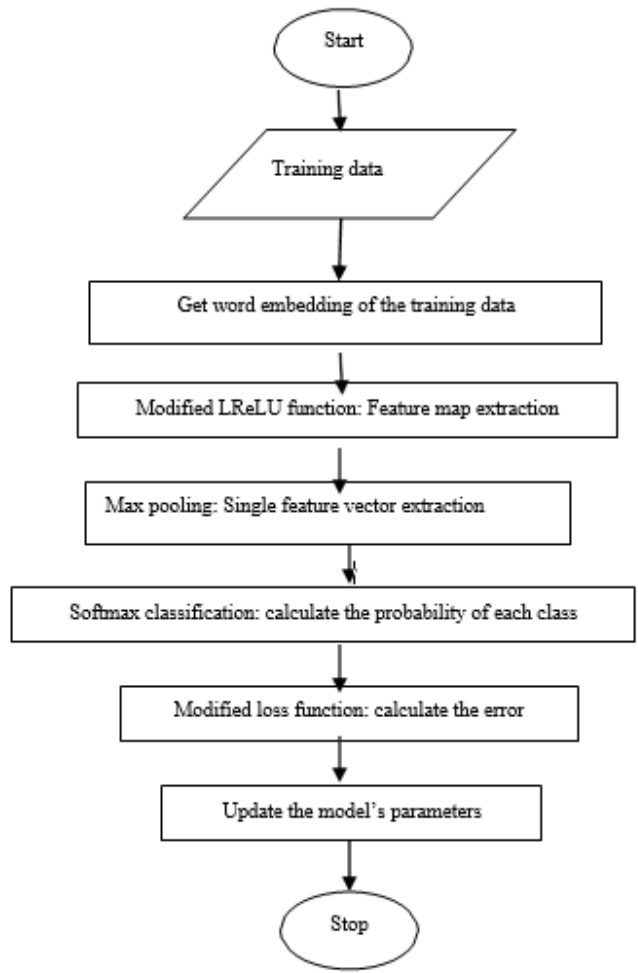

Fig. 4: Flowchart of proposed Enhanced Leaky Rectified Linear Unit activation and Weighted Loss (ELReLUWL) for text emotion classification.

## 3.1 Proposed Equation

The proposed solution replaces the sigmoid function with a modified activation function to reduce gradient saturation, which would otherwise impact the classification performance. It uses the leaky rectified linear unit (LReLU) function as the basis, as the gradient of LReLU will never saturate. The LReLU function is expressed in equation (4) adapted from Phillips et al. [20]:

$$h(x) = \begin{cases} x, & x > 0 \\ 0.01 * x, & x \leq 0 \end{cases} \tag{4}$$

where,
$h(x)$: the activation function
x: the input
0.01: the constant slope for the negative inputs

However, the slope and the inflection point in equation (4) are constant, which is not the best parameter for the function [48]. To solve this, we also adopt the idea from the displaced rectified linear unit (DReLU) function [22], which defines a better inflection point rather than the origin. The DReLU function is expressed in equation (5):

$$d(x) = \begin{cases} x, & x > -a \\ -a, & x \leq -a \end{cases} \tag{5}$$

Where,
$d(x)$: the activation function
x: the input
a: defines the point where the inflection happens

But the gradient of equation (5) for the inputs that are on the left side of the inflection point is 0, which is undesirable. The slope definition in equation (4) can solve this problem. Therefore, equation (1) will be modified by combining the advantages of the equation (3) and (4) to propose a modified version of LReLU. The proposed solution uses this modified LReLU function as the activation function, which is expressed in equation (6):

$$Mu_j = \begin{cases} x, & x > -a \\ -ax, & x \leq -a \end{cases} \tag{6}$$

Where,
$Mu_j$: the modified activation function
x: the input
a: defines the point where the inflection happens and the slope for the input values smaller than the inflection point

The proposed solution further enhances the accuracy with a modified loss function. It combines the cross-entropy loss function with each data sample's weight to reduce the bias towards the frequent classes in the training data. More specifically, it multiplies each sample's error with a corresponding weight. The modified loss function is expressed in equation (7).

 The weight for each class is calculated with the number of samples and the number of classes, as well as the number of samples belonging to that class that depends on its ground truth label [24]. The calculation of the weight for each class is expressed in equation (8).

$$\text{ML} = -\sum_k \sum_n W_m d_k \log P_k \tag{7}$$

Where,

ML: the modified loss

k: the number of classes

n: the number of samples

m: the $m^{th}$ training sample

$d_k$: the desired output of class $k$

$P_k$: the predicted probability of class $k$

$W_m$: the weight for the $m^{th}$ training sample, and is calculated by equation (8) [49]:

$$W_m = \frac{n}{k \sum_j 1\{y_j = y_m\}} \qquad (8)$$

Where,

m: the $m^{th}$ training sample

n: the number of samples

k: the total number of classes

j: the $j^{th}$ training sample

$1\{y_j = y_m\}$ is the indicator that the $j^{th}$ sample shares the same ground truth label with the $m^{th}$ sample

Applying equation (8) to equation (7), the modified loss function can be expressed in equation (9):

$$ML = -\sum_k \sum_n \frac{n}{k \sum_j 1\{y_j = y_m\}} d_k \log P_k \qquad (9)$$

Where,

ML: the modified loss

k: the number of classes

n: the number of samples

j: the $j^{th}$ training sample

m: the $m^{th}$ training sample

$1\{y_j = y_m\}$: the indicator that the $j^{th}$ sample shares the same ground truth label with the $m^{th}$ sample

$d_k$: the desired output of class $k$

$P_k$: the predicted probability of class $k$

Thus, the enhanced overall classification performance can be expressed in equation (10):

$$EC = Mu_j + ML \qquad (10)$$

Where,

EC: the enhanced overall classification performance

$Mu_j$: the modified activation function

ML: the modified loss function

Accuracy

$$Accuracy = \frac{True\ Positives + True\ Negatives}{All\ samples} \qquad (11)$$

Where,

True Positives: the number of correctly identified positive samples

True Negatives: the number of correctly identified negative samples [30].

## 3.2 Area of Improvement

Two equations were proposed to improve the performance of the neural network. First, the modified activation function to calculate the output of neurons, as shown in equation (6). It uses the modified LReLU (MLReLU) function as the activation function instead of the sigmoid function. The purpose is to prevent the gradient saturation problem, which would impact the model's recognition ability. With the help of MLReLU function, the model is simpler and more powerful in terms of recognizing the unseen data. It, thus, has higher average class accuracy and faster convergence speed. In addition, the MLReLU function enhances performance by defining the most appropriate point where the inflection should happen. Second, the cross-entropy loss function is combined with the weight of each data point. It multiplies the error with each data point's weight to counterbalance the negative effect of uneven emotion class distribution in the training data. With the help of the proposed equation (9), it reduces the model's risk inclining to more common classes and provides more accurate classification. Hyperparameter validation experiments with deferent input values were conducted to define the best parameter value of MLReLU. The five experiments' mean value was calculated, and 0.03 was selected as the best parameter value.

## 3.3 Why Enhanced Leaky Rectified Linear Unit activation and Weighted Loss (ELReLUWL) algorithm?

Using a modified LReLU function as the activation function in CNN effectively avoids the gradient saturation problem. It uses the identity for input values larger than the inflection point, and non-zero slope for the remaining inputs. The proposed system can effectively solve the gradient saturation problem as the MLReLU function's gradient is always greater than 0. With the definition of an appropriate inflection point, it further enhances performance. In addition, the MLReLU function is simple and fast to compute, resulting in a more efficient training process. Moreover, the proposed modified loss function takes the imbalanced class distribution in the training data into account. It multiplies the error with weight to alleviate the negative effect of imbalance. Each data's weight is determined by its ground truth label and is inversely proportional to the number of samples sharing the same class. Preventing the gradient from saturating in the neural network can improve the model's average class accuracy and training speed. Due to gradient saturation, the neurons backpropagate the gradient close to zero. When the gradient becomes too small, the neural network's early layers take a long time to train. It results in the latter layers losing important information from early layers when learning, which reduces the overall classification performance. It is also difficult for the neurons to update the parameters to tune the model as it learns. This phenomenon causes poor recognition performance and slow parameter convergence. Therefore, the proposed solution that solves the gradient saturation problem enhances the average class accuracy and accelerates the parameter convergence. In addition, highly imbalanced data is commonly involved in emotion analysis. Without a suitable strategy to deal with the imbalanced data, the model will generate results like a majority vote. The introduction of the weight in the loss function effectively prevents this classification bias towards more frequent classes. The activation function used in the baseline SOTA system is the sigmoid function which is prone to suffer gradient saturation problems. The proposed system solves this problem with the modified LReLU function, which its gradient does not saturate. In other words, the proposed system has a higher classification ability and takes less time to train. Furthermore, the proposed system deals with the imbalanced training data by multiplying errors with corresponding weights. The modified loss function can effectively reduce bias and enhance average class accuracy.

## 4. Results

Python 3.6.4, along with libraries such as Keras, Tensorflow, NLTK, and matplotlib [6], were used to implement this research. Two freely available datasets of online reviews and can download from [50], the Polarity dataset [51] and the IMDB movie review dataset [35] were used. For experimental purposes, we divided the IMDB dataset into two different datasets, i.e., IMDB dataset 1 and IMDB dataset 2 to allow testing of results from datasets with and without imbalance. Therefore, three datasets in total were used, and they vary in the number of samples and data balance degree, properties of which are listed in Tables 3 and 4. Five-fold cross-validation was used, where the data were randomly divided into five equal-sized sets, of which one was used for testing and the remainder for training and validation [34]. We opted for a high number of training data due to the hyperparameters' complexity in the algorithm [52] and because many researchers used an 80/20 split when k=5.

The model was trained using stochastic gradient descent with a dropout rate of 0.4, a learning rate of 0.2, a filter size of 3, 4, 5, respectively. Batch size of 100 and a hidden unit of 200 were used. For the experiment, a system configured with Intel® Core™ i5-3337U CPU@ 1.80 GHz and 4.00GB RAM was used. The experiment is divided into two parts. The first part covers three scenarios, i.e., a small dataset where the number of samples is relatively small (Polarity dataset), a large dataset where the number of samples is relatively large (IMDB dataset 1), and an unbalanced dataset where the number of negative review samples is twice that of positive review samples (IMDB dataset 2). From each dataset, 20% of the data was used for validation, for which results are shown in Table 5. The second part was based on two classes, i.e., negative and positive reviews. A comprehensive experiment has been performed on the negative and positive review samples with results shown in Tables 6, 7, 8, 9, 10, and 11.

For the three different datasets, Metric() method from Python Keras library was used to compute accuracy and processing time. The average accuracy and processing time are calculated by using the Mean() method from the Python Numpy library. The results for the different datasets are shown in Figures 7 and 8. For the different review types, the Predict() function of the Python Keras library was used to obtain the average class accuracy. In contrast, the Now() function of the DateTime library was used to calculate the processing time. The average accuracy and processing time of the test samples of two different types of reviews were calculated using the AVERAGE() function in Microsoft Excel. The results for the various review types are compared in Figures 9, 10, 11, 12, 13, and 14.

Samples of text processed within the IMDB datasets highlighting the complexity of such reviews:

Positive: "Almost missed it. While visiting friends in Philadelphia sometime in the early 1980s, I was channel surfing after everyone else went to bed. It wasn't just Bogart I was obsessed with; but rather the entire era of those old flicks those of my age know so well. Add to that a plot like The Maltese Falcon - where so many different characters were interacting with Sacchi - and you have a piece of art as far as I'm concerned. About ten years later it appeared on TV, and I taped it."

This message is largely neutral interspersed with positive markers.

Negative: "Seagal needs to get back to basics breaking bones and kicking butt. No more of this slow-motion crap like a foreigner and in the shadows fighting like half past dead. Exit wounds showed more of his fighting skills with some wires which were ok, but then he went back to be a movie director".

This message is more complex as it contains an element that commends Segal on his fighting skills. However, the overall tone is one of dissatisfaction and disappointment. For such types of comments, fine-grained analysis achieves more accurate results.

Table 3: Statistics of the datasets

| Dataset Name | Total number of samples | Max Length (words) | Min Length (words) | Average Length (words) |
|---|---|---|---|---|
| Polarity dataset | 2000 | 2287 | 16 | 623 |
| IMDB dataset 1 | 10000 | 2385 | 11 | 224 |
| IMDB dataset 2 | 3000 | 981 | 8 | 216 |

Table 4: Emotion class distribution in the datasets

| Dataset Name | Class | Number of samples |
|---|---|---|
| Polarity dataset | Positive | 1000 |
| | Negative | 1000 |
| IMDB dataset 1 | Positive | 5000 |
| | Negative | 5000 |
| IMDB dataset 2 | Positive | 1000 |
| | Negative | 2000 |

In the feature extraction and classification stage, the CNN model was created to extract features from the labelled training data automatically. The features form feature maps which are fed into the max-pooling layer. The fully connected layer then uses the pooling layer's output to generate the emotion classification (i.e. positive or negative) with the help of the 'softmax' function. After being trained, the model is evaluated using the validation data. Fig.5 presents accuracy performance on the IMDB dataset 1 at the training stage for both the baseline SOTA solution [21] and our proposed solution. The two solutions achieve similar accuracy; however, the proposed solution takes fewer epochs to achieve optimal accuracy. This means that the proposed solution has reduced processing time for training the model for the large dataset.

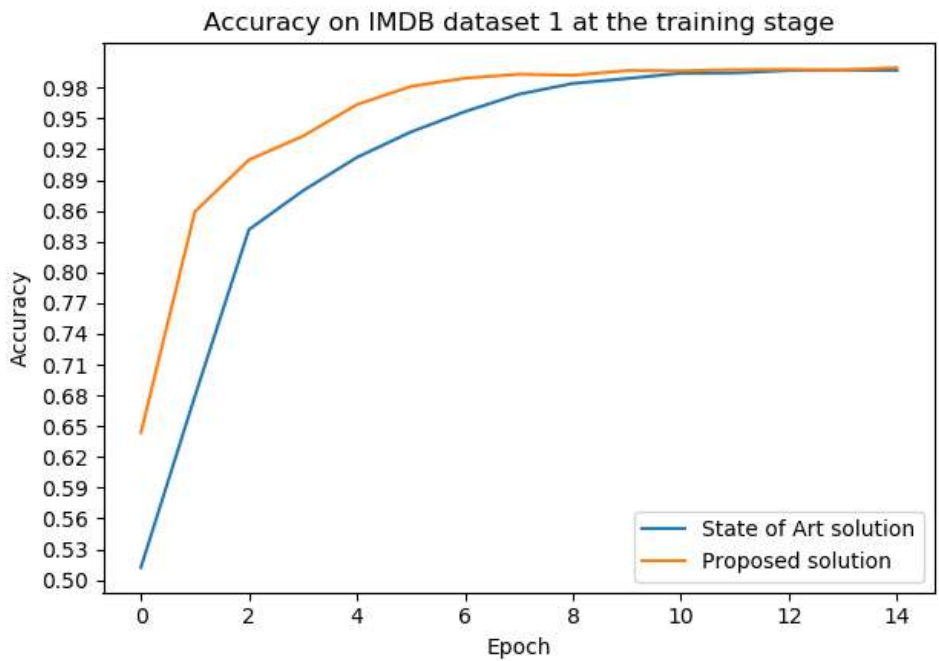

Fig.5: The average class accuracy performance on IMDB dataset 1 for baseline SOTA solution [21] and proposed solution at the training stage. a) The blue line shows the average class accuracy versus epoch of the baseline SOTA solution. b) The orange line shows the average class accuracy versus epoch of our proposed solution.

Fig.6 presents the accuracy performance on the IMDB dataset 1 at the validation stage for both the baseline SOTA solution [21] and our proposed solution. It shows that our proposed solution has dramatically improved the average class accuracy by about 5% compared to the baseline SOTA solution. Based on the training and validation stages, the classification results on the 3 different datasets are presented in Table 5, in terms of accuracy and processing time.

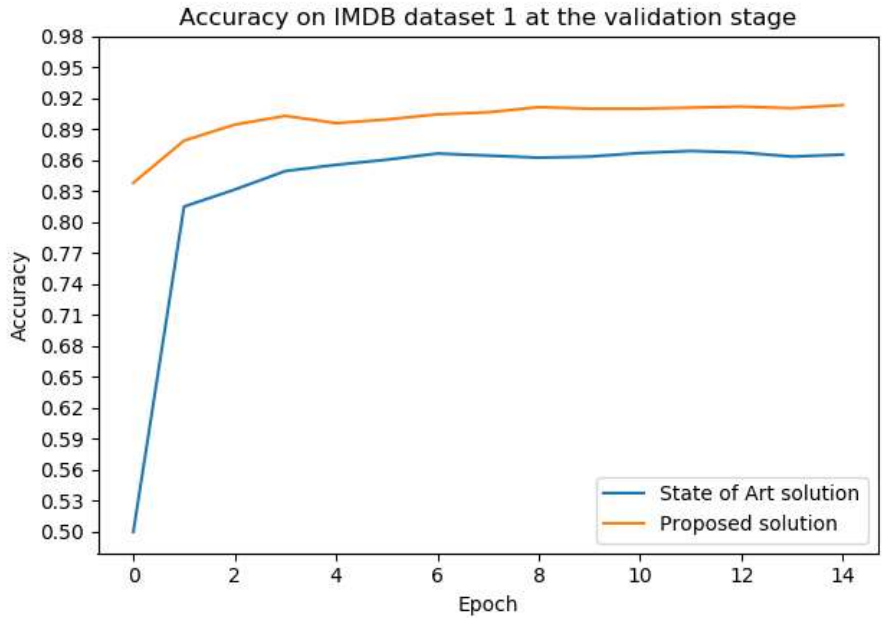

Fig.6: The average class accuracy performance on IMDB dataset 1 for baseline SOTA solution [21] and proposed solution at the validation stage. a) The blue line shows the average class accuracy versus epoch of the baseline SOTA solution. b) The orange line shows the average class accuracy versus epoch of our proposed solution.

Our proposed solution results are compared with the baseline SOTA solution using data reporting and bar graphs. The results are obtained for two aspects: 1) based on the three scenarios (i.e. small dataset, large dataset, and unbalanced dataset); 2) based on the two different review types (i.e. negative and positive). Firstly, for the different scenarios, the results are divided into three categories. For each scenario, the results are shown based on two different stages, i.e. training and validation. The results for each dataset are presented in terms of accuracy and processing time. More specifically, accuracy is calculated in terms of the percentage of correctly classified samples against the total labelled samples. And the processing time is measured in terms of the number of epochs required to reach convergence. The comprehensive test was performed on 20% of the three datasets' samples (i.e. Polarity, IMDB 1, and IMDB 2). The overall average of accuracy and processing time was calculated by averaging the training stage and validation stage results for the respective datasets, which are shown in Fig.7 and 8. For baseline SOTA, the average accuracy and processing time for the Polarity dataset is 83% and 15.5 epochs, the IMDB dataset 1 88.25% and 7.5 epochs, and the IMDB dataset 2 92.42% and 11.5 epochs. For the proposed solution, the average accuracy and processing time for the Polarity dataset is 90.85% and 10 epochs, the IMDB dataset 1 90.33% and 4 epochs and the IMDB dataset 2 95.31% and 7 epochs. Secondly, the negative and positive review samples from each dataset were tested, and the results for each sample are shown in Tables 6, 7, 8, 9, 10, and 11. We have performed the tests dividing each type (i.e. positive and negative) into ten strata to obtain average class accuracy and processing time. For the polarity dataset, each of these 20 strata consisted of 20 individual samples (total 400), while for IMDB1 and IMDB2, 20 strata of 100 (2000) and 30 (600) were used, respectively.

The average class accuracy is measured in terms of the probability of each class's labelled data while the processing time is measured in terms of the execution time to classify each test sample.

The results were obtained in the classification stage of CNN. The results are analyzed based on the two aspects mentioned above. The proposed solution has improved the average class accuracy and processing time by using equation (6) to reduce the risk of gradient saturation. And the accuracy is further enhanced with the help of equation (9) to reduce the bias in the dataset. This will improve the automatic analysis of product and service reviews for business organizations.

Table 5: Accuracy and Processing Time Results for three online review datasets

| No. | Dataset Name | Scenario | Stage | Baseline SOTA | | Proposed solution | |
|---|---|---|---|---|---|---|---|
| | | | | Accuracy (%) | Processing Time (epoch) | Accuracy (%) | Processing time (epoch) |
| 1 | Polarity | Small scaled data | Training | 88% | 19 epochs | 97.69% | 12 epochs |
| | | | Validation | 78% | 12 epochs | 84% | 8 epochs |
| 2 | IMDB 1 | Large scaled data | Training | 90.96% | 11 epochs | 91.2% | 6 epochs |
| | | | Validation | 85.55% | 4 epochs | 89.45% | 2 epochs |
| 3 | IMDB 2 | Unbalanced data | Training | 95.83% | 14 epochs | 98.04% | 8 epochs |
| | | | Validation | 89% | 9 epochs | 92.57% | 6 epochs |

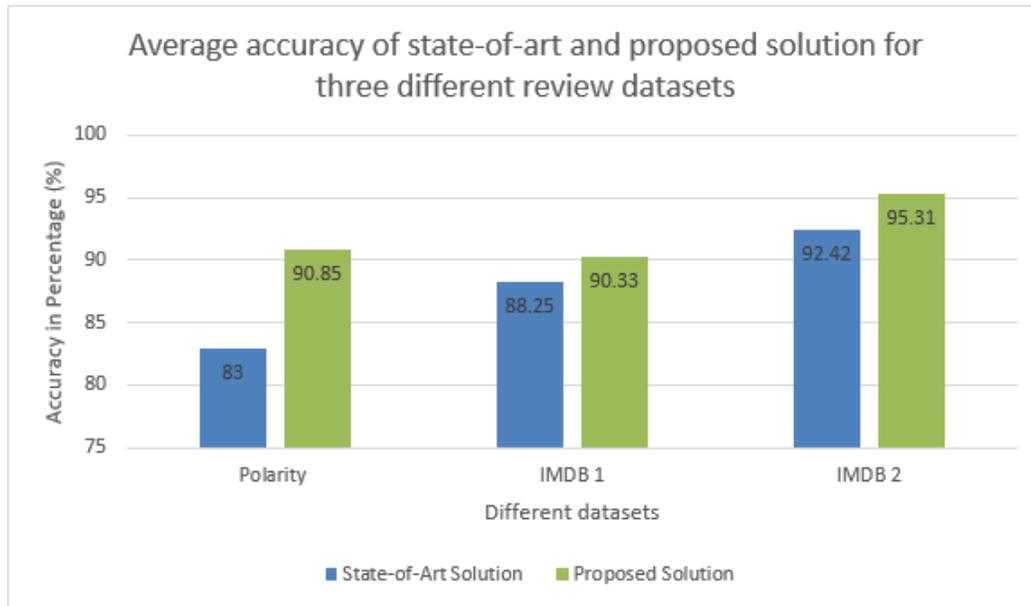

Fig.7: The average of average class accuracy in percentage for the three different review datasets. The blue color indicates the accuracy of the baseline SOTA solution [21] while the green color indicates the accuracy of the proposed solution. a) The first couple of bar graphs indicate the average accuracy for Polarity dataset. b) The second couple of bar graphs indicate the average accuracy for IMDB dataset 1. c) The third couple of bar graphs indicate the average accuracy for IMDB dataset 2

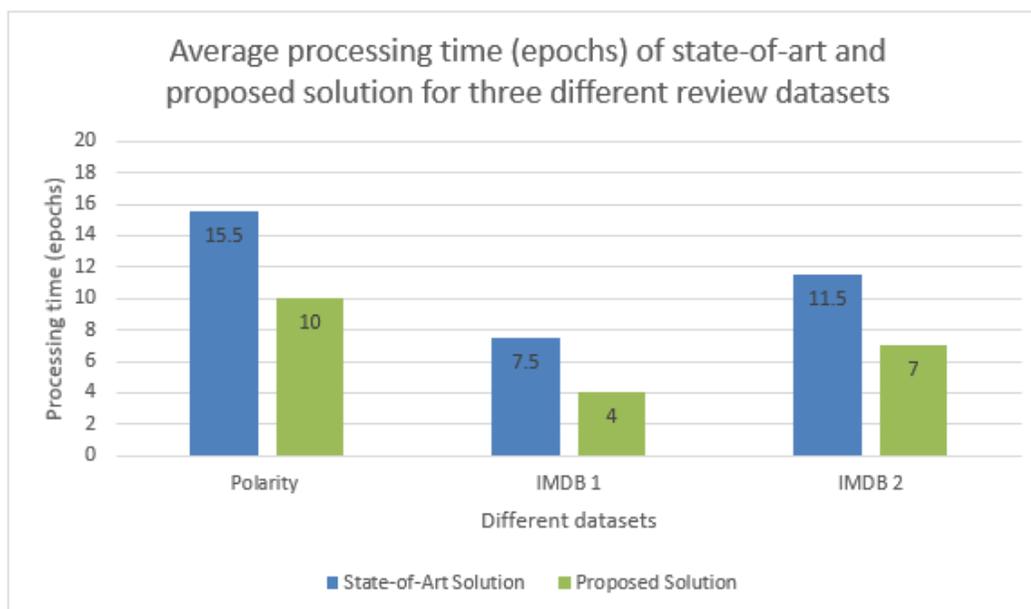

Fig.8: Average processing time in epochs for the three different review datasets. The blue color indicates the processing time of the baseline SOTA solution [21] while the green color indicates the processing time of the proposed solution. a) The first couple of bar graphs indicate the average processing time for Polarity dataset. b) The second couple of bar graphs indicate the average processing time for IMDB dataset 1. c) The third couple of bar graphs indicate the average processing time for IMDB dataset 2.

Table 6: Accuracy and Processing time Results of Baseline SOTA solution [21] and Proposed solution for negative review samples from Polarity dataset

| Sample No. | Sample group details | Baseline SOTA solution | | Proposed solution | |
|---|---|---|---|---|---|
| | | Accuracy (%) | Processing Time (Millisecond) | Accuracy (%) | Processing time (Millisecond) |
| 1.1 | Negative review samples from Polarity dataset | 91.27% | 29.8ms | 97.07% | 19.9ms |
| 1.2 | | 87.36% | 29.7ms | 95.52% | 19.8ms |
| 1.3 | | 90.64% | 29.9ms | 95.85% | 19.9ms |
| 1.4 | | 91% | 29.9ms | 97.02% | 19.9ms |
| 1.5 | | 87.86% | 29.8ms | 94.10% | 19.8ms |
| 1.6 | | 91.06% | 29.9ms | 96.71% | 19.7ms |
| 1.7 | | 93.34% | 30.1ms | 98.13% | 19.9ms |
| 1.8 | | 89.81% | 29.2ms | 94.78% | 20ms |
| 1.9 | | 87.06% | 29.9ms | 93.62% | 20ms |
| 1.10 | | 89.58% | 29.9ms | 96.03% | 20.2ms |

Table 7: Accuracy and Processing time Results of Baseline SOTA solution [21] and Proposed solution for positive review samples from Polarity dataset

| Sample No. | Sample group details | Baseline SOTA solution | | Proposed solution | |
|---|---|---|---|---|---|
| | | Accuracy (%) | Processing Time (Millisecond) | Accuracy (%) | Processing time (Millisecond) |
| 2.1 | Positive review samples from Polarity dataset | 93.95% | 39.7ms | 99.41% | 30.3ms |
| 2.2 | | 90.53% | 39.7ms | 95.35% | 29.9ms |
| 2.3 | | 92.34% | 39.6ms | 98.61% | 29.9ms |
| 2.4 | | 88.44% | 40ms | 94.25% | 29.9ms |
| 2.5 | | 91.9% | 39.9ms | 96.77% | 29.7ms |
| 2.6 | | 87.98% | 39.9ms | 93.75% | 29.9ms |
| 2.7 | | 91.13% | 39.8ms | 95.11% | 29.5ms |
| 2.8 | | 91.84% | 39.5ms | 96.83% | 30ms |
| 2.9 | | 93.62% | 39.2ms | 99.18% | 29.6ms |
| 2.10 | | 87.98% | 39.5ms | 95.7% | 29.6ms |

Table 8: Accuracy and Processing time Results of Baseline SOTA solution [21] and Proposed solution for negative review samples from IMDB dataset 1

| Sample No. | Sample group details | Baseline SOTA solution | | Proposed solution | |
|---|---|---|---|---|---|
| | | Accuracy (%) | Processing Time (Millisecond) | Accuracy (%) | Processing time (Millisecond) |
| 3.1 | Negative review samples from IMDB dataset 1 | 96.97% | 29.9ms | 99.76% | 19.8ms |
| 3.2 | | 88.16% | 29.9ms | 92.9% | 20.1ms |
| 3.3 | | 92.64% | 30ms | 99.99% | 19.9ms |
| 3.4 | | 96.99% | 29.9ms | 99.58% | 19.7ms |
| 3.5 | | 96.34% | 29.6ms | 99.76% | 20ms |
| 3.6 | | 93.9% | 29.9ms | 96.89% | 20.2ms |
| 3.7 | | 94.72% | 29.5ms | 97.81% | 19.9ms |
| 3.8 | | 95.83% | 30ms | 98.59% | 19.9ms |
| 3.9 | | 89.42% | 29.9ms | 94.18% | 20ms |
| 3.10 | | 89.66% | 30.1ms | 93.73% | 19.9ms |

Table 9: Accuracy and Processing time Results of Baseline SOTA solution [21] and Proposed solution for positive review samples from IMDB dataset 1.

| Sample No. | Sample group details | Baseline SOTA solution | | Proposed solution | |
|---|---|---|---|---|---|
| | | Accuracy (%) | Processing Time (Millisecond) | Accuracy (%) | Processing time (Millisecond) |
| 4.1 | Positive review samples from IMDB dataset 1 | 91.85% | 29.8ms | 95.55% | 19.9ms |
| 4.2 | | 96.08% | 29.6ms | 99.83% | 19.8ms |
| 4.3 | | 90.35% | 30ms | 94.14% | 19.9ms |
| 4.4 | | 96.01% | 30.1ms | 99.17% | 20ms |
| 4.5 | | 92.48% | 29.9ms | 96.92% | 20.1ms |
| 4.6 | | 94.91% | 30ms | 98.29% | 20.2ms |
| 4.7 | | 94.42% | 29.9ms | 98.71% | 19.9ms |
| 4.8 | | 91.93% | 30.2ms | 95.04% | 20.2ms |
| 4.9 | | 92.88% | 29.7ms | 96.95% | 20ms |
| 4.10 | | 95.87% | 30ms | 99.23% | 19.9ms |

Table 10: Accuracy and Processing time Results of Baseline SOTA solution [21] and Proposed solution for negative review samples from IMDB dataset 2.

| Sample No. | Sample group details | Baseline SOTA solution | | Proposed solution | |
|---|---|---|---|---|---|
| | | Accuracy (%) | Processing Time (Millisecond) | Accuracy (%) | Processing time (Millisecond) |
| 5.1 | Negative review samples from IMDB dataset 2 | 89.39% | 40.2ms | 95.4% | 30ms |
| 5.2 | | 93.85% | 39.9ms | 99.03% | 29.9ms |
| 5.3 | | 94.77% | 40ms | 99.22% | 29.9ms |
| 5.4 | | 89.36% | 39.9ms | 94.75% | 30ms |
| 5.5 | | 94.86% | 39.9ms | 98.98% | 29.8ms |
| 5.6 | | 92.26% | 40ms | 97.92% | 30.2ms |
| 5.7 | | 90.24% | 39.9ms | 96.59% | 29.9ms |
| 5.8 | | 95.81% | 39.9ms | 99.99% | 30ms |
| 5.9 | | 88.06% | 40ms | 94.13% | 29.8ms |
| 5.10 | | 88.68% | 39.8ms | 94.32% | 30.1ms |

Table 11: Accuracy and Processing time Results of Baseline SOTA solution [21] and Proposed solution for positive review samples from IMDB dataset 2.

| Sample No. | Sample group details | Baseline SOTA solution | | Proposed solution | |
|---|---|---|---|---|---|
| | | Accuracy (%) | Processing Time (Millisecond) | Accuracy (%) | Processing time (Millisecond) |
| 6.1 | Positive review samples from IMDB dataset 2 | 93.95% | 39.7ms | 99.41% | 30.3ms |
| 6.2 | | 90.53% | 39.7ms | 95.35% | 29.9ms |
| 6.3 | | 92.34% | 39.6ms | 98.61% | 29.9ms |
| 6.4 | | 88.44% | 40ms | 94.25% | 29.9ms |
| 6.5 | | 91.9% | 39.9ms | 96.77% | 29.7ms |
| 6.6 | | 87.98% | 39.9ms | 93.75% | 29.9ms |
| 6.7 | | 91.13% | 39.8ms | 95.11% | 29.5ms |
| 6.8 | | 91.84% | 39.5ms | 96.83% | 30ms |
| 6.9 | | 93.62% | 39.2ms | 99.18% | 29.6ms |
| 6.10 | | 87.98% | 39.5ms | 95.7% | 29.6ms |

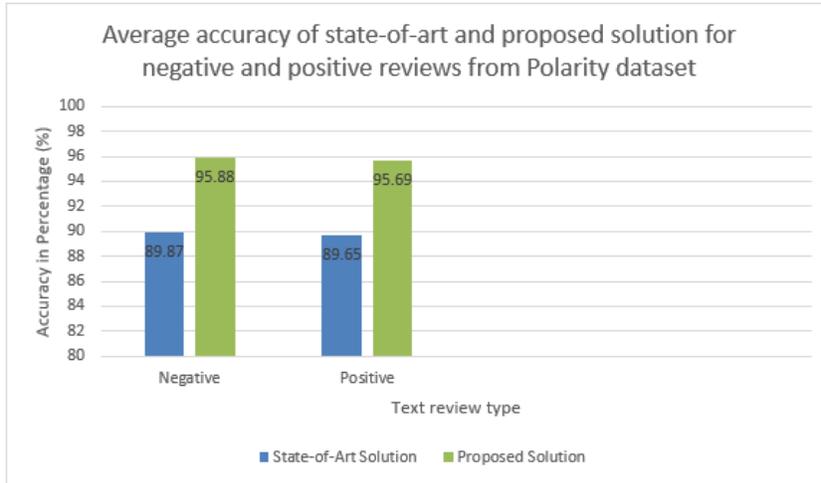

Fig.9: Average class accuracy in percentage for Negative and Positive reviews from Polarity dataset. The blue color indicates the accuracy of the baseline SOTA solution [21] while the green color indicates the accuracy of the proposed solution. a) The first couple of bar graphs indicate the average accuracy for Negative reviews. b) The second couple of bar graphs indicate the average accuracy for Positive reviews.

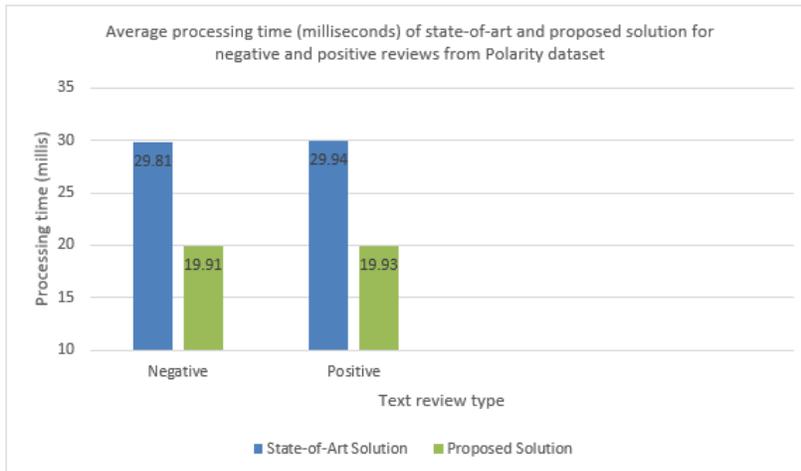

Fig.10: Average processing time in milliseconds for Negative and Positive reviews from Polarity dataset. The blue color indicates the processing time of the baseline SOTA solution [21] while the green color indicates the processing time of the proposed solution. a) The first couple of bar graphs indicate the average processing time for Negative reviews. b) The second couple of bar graphs indicate the average processing time for Positive reviews.

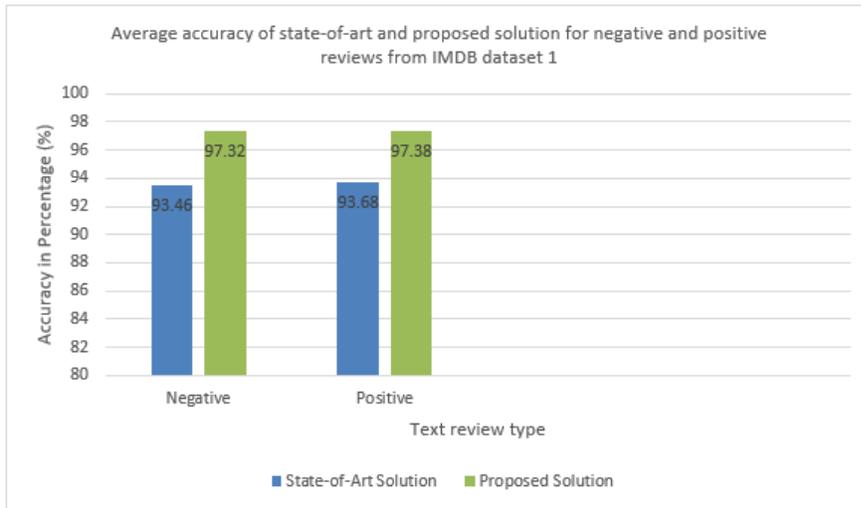

Fig.11: The average of average class accuracy in percentage for Negative and Positive reviews from IMDB dataset 1. The blue color indicates the accuracy of the baseline SOTA solution [21] while the green color indicates the accuracy of the proposed solution. a) The first two bar graphs indicate the average accuracy for Negative reviews. b) The second couple of bar graphs indicate the average accuracy for Positive reviews.

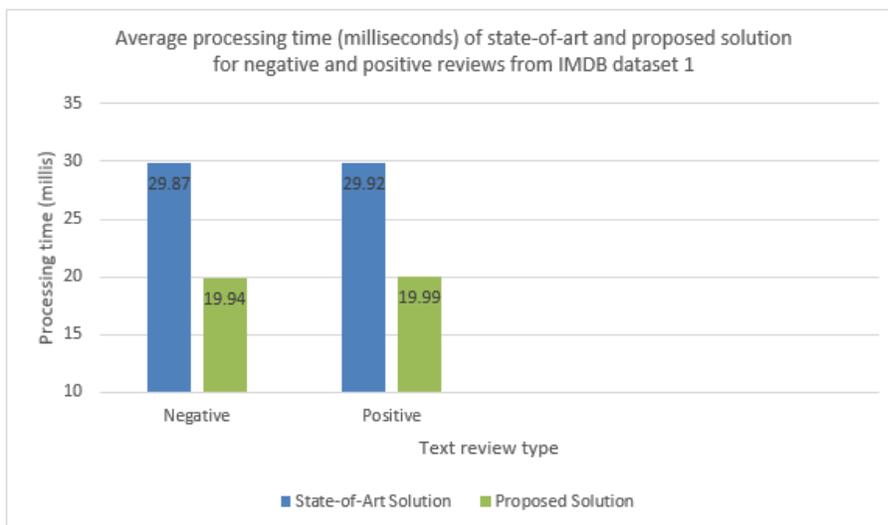

Fig.12: Average processing time in milliseconds for Negative and Positive reviews from IMDB dataset 1. The blue color indicates the processing time of the baseline SOTA solution [21] while the green color indicates the processing time of the proposed solution. a) The first couple of bar graphs indicate the average processing time for Negative reviews. b) The second couple of bar graphs indicate the average processing time for Positive reviews.

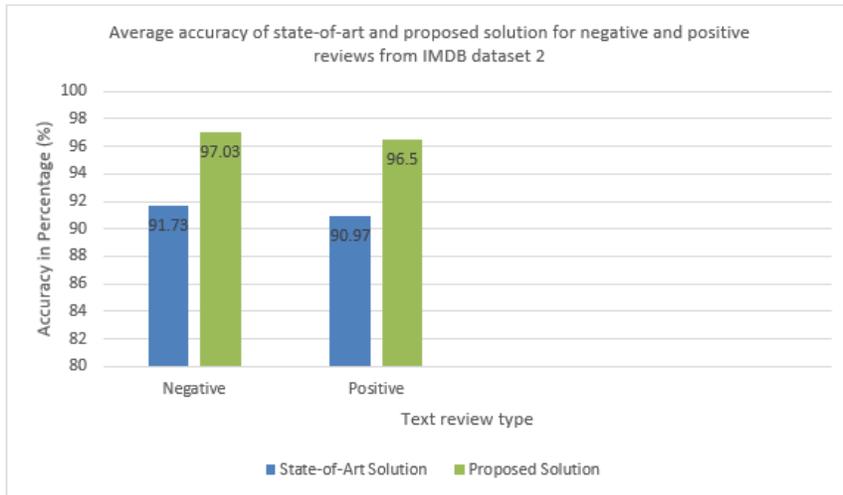

Fig.13: Average The average of average accuracy in percentage for Negative and Positive reviews from IMDB dataset 2. The blue color indicates the accuracy of the baseline SOTA solution [21] while the green color indicates the accuracy of the proposed solution. a) The first couple of bar graphs indicate the average accuracy for Negative reviews. b) The second couple of bar graphs indicate the average accuracy for Positive reviews.

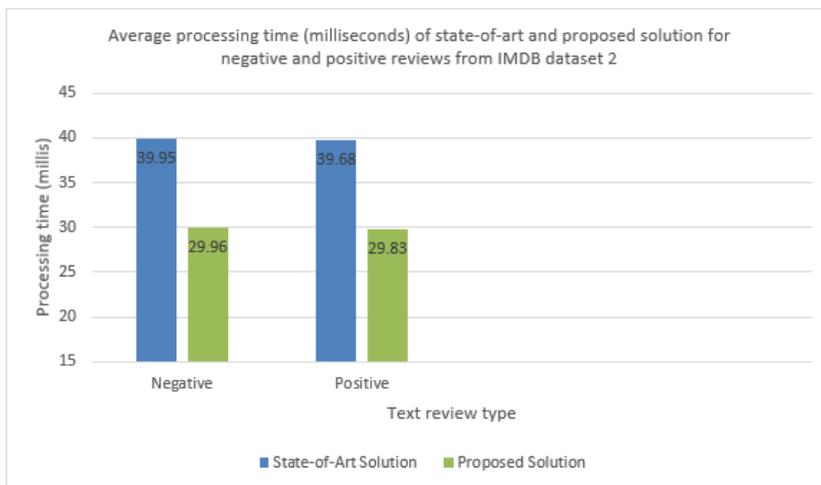

Fig.14: Average processing time in milliseconds for Negative and Positive reviews from IMDB dataset 2. The blue color indicates the processing time of the baseline SOTA solution [21] while the green color indicates the processing time of the proposed solution. a) The first couple of bar graphs indicate the average processing time for Negative reviews. b) The second couple of bar graphs indicate the average processing time for Positive reviews.

## 5. Discussion

The results show the improvement in average class accuracy and processing time of the proposed solution compared to the baseline SOTA solution based on text emotion classification. With the modified activation function and modified loss function, the proposed solution provides an average class accuracy of 96.63%, which is more than 5% higher than the current solution [21]. The proposed model also requires less convergence time to achieve optimization with 7 epochs which are 4.5 epochs below the requirements of the current solution.

In addition, the proposed solution decreases the average processing time to 23.3 milliseconds which is 10 milliseconds less below that needed by the current solution. The accuracy for each sample is obtained using the Predict() method from the Python Keras package where true positives and true negatives are used. Processing time is obtained by using the Now() method of the Python DateTime package where the start and stop times are used. The average accuracy and processing time are calculated using the Average() function in Microsoft Excel. The overall degree of improvement in average class accuracy and processing time is quantified by running the baseline SOTA algorithm and the proposed solution algorithm, respectively. Accuracy is calculated using equation (11) above [30].

Using the modified Leaky ReLU (MLReLU) function as the activation function in the convolutional layer prevents gradient saturation and improves the proposed system's performance. This modified feature is inspired by the work of Phillips et al. [20]. The main benefit of the MLReLU function is that its gradient will never saturate so that the parameters can be updated effectively to achieve model optimization. The use of a weighted loss function is another feature of our proposed solution, which was adapted from Kratzwald et al. [24]. It reduces the bias in the imbalanced data and provides enhanced accuracy. Thus, the proposed solution provides improved average class accuracy and reduced processing time when compared to the baseline SOTA solution. In conclusion, CNN's combination with the proposed ELReLUWL algorithm has greatly improved the online review emotion classification with enhanced accuracy of 96.63% and reduced processing time of 23.3 milliseconds.

A significant number of techniques has been implemented for emotion classification of text. These techniques' goals are consistent: to achieve higher average class accuracy while using lower processing time. The current solution's limitations have been successfully overcome with this research, with an average accuracy of 96.63% against the current accuracy of 91.56%. This research also reduces the average processing time to 23.3 milliseconds against the current processing time of 33.2 milliseconds. This is due to improvements in the activation function to minimize the risk of gradient saturation and the weighted loss function to reduce data bias. Hence, the proposed solution has better performance across different data scenarios (i.e., small dataset, large dataset, and unbalanced dataset) and different review types (i.e., positive and negative reviews).

# 6. Conclusion

Understanding emotions from online reviews is particularly important for marketers to capture user opinions and attitudes to products and services. Accurate classification of emotion from the reviews boosts the chance of business success. Many deep learning approaches have been utilized to achieve this. However, there are limitations in terms of accuracy and processing time. A modified leaky ReLU activation function has been proposed, which reduces the risk of gradient saturation when the model is trained to improve accuracy and convergence speed. The weighted loss function is another feature adapted for this research to minimise the negative effect of bias in the imbalanced dataset and further enhance average class accuracy. Therefore, the proposed solution has reduced the convergence time by 4.5 epochs on average, improved accuracy by more than 5% on average, and decreased the processing time by 10 milliseconds on average. Because there are no publicly available labelled datasets for multi-class emotion classification, this research is limited to binary classification (i.e., positive and negative). In the future study, multi-class datasets for different emotion types (e.g., joyful, surprise, angry, sad, and so on) should be used to train and test the model to extend it to broader domains. Other enhancement methods such as a data augmentation technique could be leveraged for better feature extraction and further advancing the classification performance.